\documentclass{svproc}
\usepackage{url}
\usepackage{comment}
\usepackage{caption}
\usepackage{booktabs}
\usepackage{placeins}
\usepackage{amsmath}
\usepackage{graphicx}
\usepackage{url}
\usepackage{float}

\begin{document}
\mainmatter              % start of a contribution
\title{Unleashing the Power of User Reviews: Exploring Airline Choices at Catania Airport, Italy}
%\title{The impact of user reviews on the choice of the airlines: a case study of Catania airport in Italy}
%

\titlerunning{Unleashing the Power of User Reviews}
%\titlerunning{The impact of user reviews on the choice of the airlines}  % abbreviated title (for running head)
\author{Vincenzo Miracula\inst{1} \and Antonio Picone\inst{1}}
\authorrunning{Vincenzo Miracula, Antonio Picone} % abbreviated author list (for running head)

\institute{University of Catania, Catania, CT 95123, ITALY,\\
\email{vincenzo.miracula@phd.unict.it}
\\
\email{antonio.picone@phd.unict.it}}

\maketitle              % typeset the title of the contribution

\begin{abstract}
This study aims to investigate the possible relationship between the mechanisms of social influence and the choice of airline, through the use of new tools, with the aim of understanding whether they can contribute to a better understanding of the factors influencing the decisions of consumers in the aviation sector. We have chosen to extract user reviews from well-known platforms: Trustpilot, Google, and Twitter. By combining web scraping techniques, we have been able to collect a comprehensive dataset comprising a wide range of user opinions, feedback, and ratings. We then refined the BERT model to focus on insightful sentiment in the context of airline reviews. Through our analysis, we observed an intriguing trend of average negative sentiment scores across various airlines, giving us deeper insight into the dynamics between airlines and helping us identify key partnerships, popular routes, and airlines that play a central role in the aeronautical ecosystem of Catania airport during the specified period. Our investigation led us to find that, despite an airline having received prestigious awards as a low-cost leader in Europe for two consecutive years 2021 and 2022, the "Catanese" user tends to suffer the dominant position of other companies. Understanding the impact of positive reviews and leveraging sentiment analysis can help airlines improve their reputation, attract more customers, and ultimately gain a competitive edge in the marketplace.
\end{abstract}
\section{Introduction}

In many social and biological systems, individuals rely on the observation of others to adapt their behaviors, revise their judgments, or make decisions \cite{Couzin},\cite{Bikhchandani},\cite{Moussaid}. In human populations, access to social information has been greatly facilitated by the continued growth of communication technology. Indeed, people are constantly exposed to a steady stream of others' opinions, advice, and judgments about political ideas, new technologies, or commercial products \cite{Wu}. When facing peer opinions on a given issue, people tend to filter and integrate the social information they receive and adjust their beliefs accordingly \cite{Yaniv}. At the scale of a group, repeated local influences among group members can give rise to complex patterns of opinion dynamics such as consensus formation, polarization, or fragmentation \cite{Mäs},\cite{Isenberg},\cite{Galam}. For example, it has been found that people who share similar extreme views, such as racial prejudice, tend to strengthen their judgment and trust after interacting with each other \cite{Myers}.

Similar mechanisms of opinion dynamics can take place in a variety of social contexts, such as within a group of friends exchanging opinions about their willingness to be vaccinated against the flu \cite{Funk},\cite{Funk1}. On an even larger scale, local influences among friends, family members, or colleagues, often combined with global mass media effects, constitute an effective mechanism that drives opinion formation during elections, shaping cultural markets \cite{Salganik}, producing amplification or attenuation of risk perception \cite{Kasperson}, and shaping public opinion on social issues, such as atomic energy or climate change \cite{Latane}.

Given the remarkably broad scope of social phenomena that are shaped by social influence and opinion dynamics, it is surprising that the behavioral mechanisms underlying these processes remain poorly understood. Major questions remain unanswered: how do people adjust their judgments during social interactions? What are the heuristics underlying opinion adaptation? And how do these local influences ultimately generate global patterns of opinion change? Much of the existing modeling work on opinion dynamics has been approached from a physics-based point of view, where the basic mechanisms of social influence derive from analogies with physical systems, particularly the spin systems \cite{Castellano}, \cite{Lorenz}. 

The wide variety of existing models assumes that individuals hold different rules for adapting their opinions, such as imitation \cite{Deffuant}, averaging with people with similar opinions \cite{Rainer}, or following the majority \cite{Lorenzmaggioranza}. Although informative about the complex dynamics that can potentially emerge in a collective context, these simulation-based contributions share a common drawback: the absence of empirical verification of the models' hypotheses.

Indeed, it's quite a challenge to trace and measure how opinions change under experimental conditions, as these changes depend on numerous social and psychological factors, such as individuals' personalities, confidence levels, credibility, social status, or persuasive power. In other fields like cognitive sciences and social sciences, laboratory experiments have been conducted to examine how people integrate feedback from others to revise their initial responses to factual questions \cite{Yaniv}, \cite{Lorenzmaggioranza}, \cite{Soll}. However, the outcomes of local rules for opinion adaptation have yet to be employed in studying the collective dynamics of the system, and it remains unclear how social influence unfolds over time in large-scale social contexts \cite{Mason}

This study aims to investigate the relationship between social influence mechanisms and airline choice while seeking to address some fundamental questions. What are the factors that sway individuals towards one airline over another? Does the user have real freedom to choose the airline carrier? Is there a hierarchy among airports? We specifically focus on flights departing from and arriving at Catania Airport, given the significance of this airport hub in the heart of the Mediterranean.

\section{Materials and Method}
% HO INSERITO QUALCOSA MA VA RIVISTO PER ADATTARLO AL NOSTRO LAVORO. 
%
To gather data for our analysis, we have chosen to extract user reviews from two well-known platforms: Trustpilot and Google. Trustpilot offers a platform for users to share their feedback and reviews about various businesses and services, while Google allows users to leave reviews for apps, products, and services.

By collecting reviews from Trustpilot and Google, our objective is to capture a wide spectrum of user opinions and experiences. These platforms attract a substantial user base that shares feedback on various aspects of the apps, products, or services they have utilized. Analyzing reviews from both platforms enables us to gain valuable insights into the sentiments and perspectives of users within our specific context.

In order to gather the necessary data for our analysis, we utilized Python in conjunction with the BeautifulSoup library. The process involved web scraping to extract information from Trustpilot, while an official API was employed to extract data from Google. For the scraping part, we leveraged BeautifulSoup's powerful parsing capabilities to navigate the HTML structure of Trustpilot's web pages. With BeautifulSoup, we were able to locate and extract relevant information such as user reviews, ratings, and reviewers' names. By programmatically traversing the HTML elements, we systematically collected the necessary data for further analysis. On the other hand, when it came to extracting Google reviews, we relied on an official API. They provide a structured and documented way to interact with a particular platform or service. In this case, the Google API allowed us to access and retrieve specific data related to reviews, ratings, and other relevant details directly from the Google platform. By following the API's guidelines and utilizing the appropriate endpoints, we could efficiently retrieve the desired data in a standardized format.

It's important to note that while scraping involves extracting data from the presentation layer of a web page, APIs offer a more streamlined and controlled approach to accessing data from the source directly. APIs often provide specific methods and endpoints tailored to retrieve data in a structured manner, ensuring consistent and reliable results. This eliminates the need for parsing and navigating HTML structures, simplifying the data extraction process and ensuring compatibility with the platform's terms of service. Scraping certainly requires higher skills than the use of an API and a further cleaning of the data, sometimes soiled by the presence of HTML tags that must be removed, in order to obtain text in its classic form.

There is a common phenomenon known as ``negativity bias"\cite{negativitybias} that suggests people are more inclined to share negative experiences and write negative reviews on the internet. Several factors contribute to this trend. 

Firstly, negative experiences often have a more significant emotional impact on individuals compared to positive experiences. As a result, people may feel a stronger urge to express their dissatisfaction or frustration by writing negative reviews as a means of catharsis. Secondly, negative reviews can serve as a warning to others, potentially preventing them from encountering similar negative experiences. People may feel a sense of responsibility to share their negative encounters to protect or inform others.

Thirdly, there is a perception that negative reviews have more weight and credibility. Some individuals may believe that expressing criticism or highlighting flaws will lead to improvements in the product, service, or overall customer experience. This belief can incentivize people to voice their negative opinions.

Twitter - on the other side - can indeed serve as a platform to mitigate the negativity bias issue when it comes to online reviews and feedback. The real-time nature of Twitter provides an advantage in mitigating the negativity bias issue. Users can share their thoughts and experiences on the platform as they happen, allowing for a more balanced representation of both positive and negative sentiments. This immediacy enables individuals to express their satisfaction, appreciation, or dissatisfaction with airlines or their services at the moment, capturing a diverse range of opinions and experiences. Data from this source were also extracted by using the official Twitter API.

By combining web scraping and API usage, we were able to collect a comprehensive dataset from Trustpilot, Google, and Twitter, encompassing a wide range of user opinions, feedback, and ratings. This rich dataset allows for in-depth analysis and insights into the experiences and perspectives of users within our specific research context. The dataset, consisting of the cleaned textual reviews and the metadata mentioned, was saved in serialized JSON (JavaScript Object Notation) format, for easy reading by both a human being and a computer, as further analysis of sentiment followed, obtained through the use of machine learning techniques applied to Natural Language Processing (NLP). 

To ensure accurate results, we initially performed text pre-processing, which involved cleaning and preparing the airline reviews for analysis. This step included removing unnecessary characters, punctuation, and stopwords, as well as tokenizing the text into meaningful units. Next, we employed BERT (Bidirectional Encoder Representations from Transformers)\cite{von2017advances} for sentiment analysis. BERT is a state-of-the-art NLP model that has demonstrated exceptional performance in understanding language context. Instead of training BERT from scratch, we utilized a pre-trained BERT model that had already learned from a vast amount of text data. To tailor the pre-trained BERT model specifically for sentiment analysis of airline reviews, we further fine-tuned it. Fine-tuning involved training the model on a labeled dataset that included positive and negative sentiment labels. By leveraging the pre-existing knowledge and linguistic understanding of the BERT model, we refined it to focus on discerning sentiment in the context of airline reviews.

This approach allowed us to benefit from BERT's powerful language representation capabilities and effectively adapt it to the task of sentiment analysis for airline reviews. The fine-tuned BERT model was then able to provide more accurate and reliable sentiment predictions, facilitating a comprehensive analysis of the sentiments expressed in the reviews.

The data gathered thus far, while valuable, do not encompass the entirety of the information required to conduct our analysis comprehensively. In addition to the existing data, we need to include data from airline schedules, including scheduled departures and arrivals; real departures and arrivals times along with data about the airline of the aircraft. By integrating these additional data points, we can form a more holistic view of the airline industry and the dynamics of flights to and from Catania Airport.

To retrieve and save flight data for flights to and from Catania airport, we developed Python code leveraging the ADSBexchange.com website. ADSBexchange.com provides online access to ADS-B (Automatic Dependent Surveillance-Broadcast) data, enabling precise tracking of aircraft movements, including takeoff and landing times. We utilized Python to interact with the APIs to get the required data from the website, extracting information about specific flights associated with Catania airport. The retrieved data, including flight details and timestamps, were then stored in a MongoDB database. MongoDB, a popular NoSQL database, offered a flexible and scalable solution to store and organize the flight data efficiently. By combining web scraping techniques, the ADSBexchange.com APIs, and MongoDB, we were able to gather and store accurate flight information for comprehensive analysis and further insights.

\section{Results}
For our analysis, we have chosen a specific period from February 18, 2023, to May 18, 2023, spanning three months. This timeframe allows us to capture a substantial amount of data for in-depth analysis. During this period, a total of 16,590 flights were recorded, encompassing both arrivals and departures at the airport. To gain insights into the prominent airlines operating at the airport, we extracted data on the number of flights operated by each airline to and from the airport. By evaluating the flight volume, we were able to identify the top 10 airlines (see Table 1) with the highest number of flights, providing us with valuable information on the major players in the aviation industry at Catania airport during the specified timeframe. 

\begin{table}[h]
    \caption{Flights per airline}
    \centering
    \begin{tabular}{l|l|c}
    \hline
    \textbf{\#} & \textbf{Airline} & \textbf{Flights} \\
    \hline
    1 & Ryanair & 6257 \\
    2 & ITA Airways & 2271 \\ 
    3 & Wizz Air & 2021 \\
    4 & easyJet & 1796 \\
    5 & Eurowings~ & 332 \\
    6 & Air Malta & 274 \\
    7 & Vueling & 261 \\
    8 & Poste Air Cargo  & 257 \\
    9 & Volotea & 237 \\
    10 & Lufthansa & 234 \\
    \hline
    
    \end{tabular}
    
    \label{tab:my_label}
\end{table}

It should be noted that among the top 10 airlines, "Poste Air Cargo" appears, which will not be examined because it is a flight carrying cargo and not a commercial passenger one.

\begin{figure}[H]
    \centering
    \includegraphics[width=9cm]{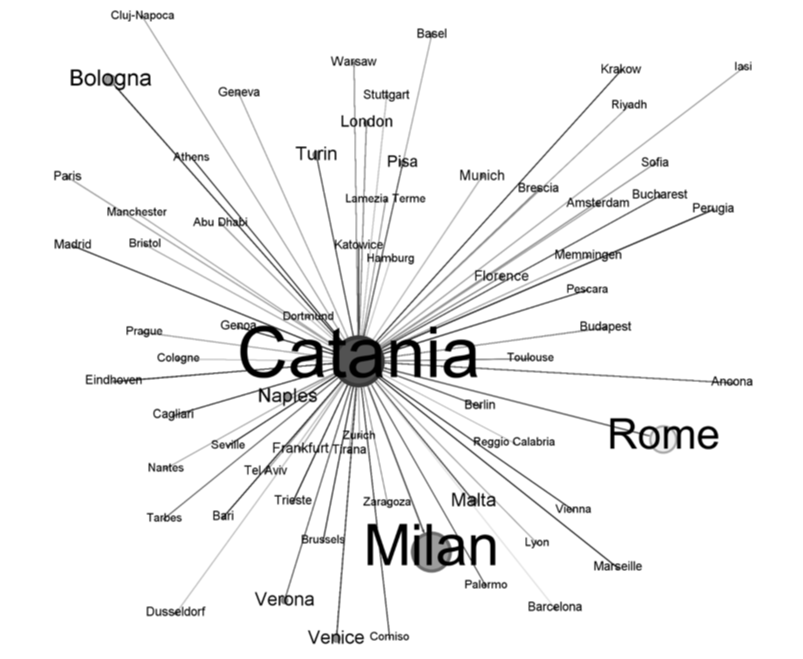}
    \caption{Flight route network Catania airport}
    \label{fig:network}
\end{figure}

In addition to this analysis, we conducted a network analysis (Figure \ref{fig:network}) to obtain a more comprehensive understanding of the airline ecosystem at Catania airport during the specified period. Using flight route data and the involved airlines, we constructed a network model that represents the connections between the airlines based on the operated flights.

Through this network analysis, we were able to identify the relationships and interactions among the airlines, including both direct and indirect connections. We evaluated the centrality of airlines in the network, considering metrics such as the degree of connectivity, the importance of connections, and the strategic position within the network.

This network analysis provided us with a deeper insight into the dynamics among the airlines and helped us identify key partnerships, the most frequent routes, and the airlines playing a central role in the aviation ecosystem at Catania airport during the specified period. This information can be valuable in understanding the competitive structure of the aviation industry and the collaboration strategies among the involved airlines.

%\noindent\begin{minipage}{\linewidth}
%\centering
%\captionof{table}{Flights per airline}\begin{tabular}{l|l|l} 
%\toprule
%\multicolumn{1}{c|}{\textbf{\#}} & \textbf{Airline} & \multicolumn{1}{c}{\textbf{Flights}}  \\ 
%\hline
%1                                & Ryanair          & 6257                           %       \\
%2                                & ITA Airways      & 2271                           %       \\
%3                                & Wizz Air         & 2021                           %       \\
%4                                & easyJet          & 1796                           %       \\
%5                                & Eurowings~       & 332                            %       \\
%6                                & Air Malta        & 274                            %       \\
%7                                & Vueling          & 261                            %       \\
%8                                & Poste Air Cargo  & 257                            %       \\
%9                                & Volotea          & 237                            %       \\
%10                               & Lufthansa        & 234                            %       \\
%\bottomrule
%\end{tabular}
%\end{minipage}

\subsection{Sentiment analysis for top 10 company}

The collection of text data starts from the date of 26 April 2022, chosen as the starting point for the collection of reviews because it corresponds to the date of the first available review for ITA Airways - which is the newest airline in the list. By starting data collection from this date, we ensure that our data set remains balanced and includes revisions for the same period. This approach allows us to maintain fairness and completeness in our analysis by including an equal representation of customer sentiment across different airlines over time (see Table 2).
\\

\noindent\begin{minipage}{\linewidth}
\centering
\captionof{table}{Sentiment Analysis on airline's review}\begin{tabular}{c|c|c} 
\toprule
\textbf{Airline} & \textbf{Positive} & \textbf{Negative}  \\ 
\hline
Ryanair          & 10.56\%           & 89.44\%            \\
ITA Airways      & 8.94\%            & 91.06\%            \\
Wizz Air         & 3.6\%             & 96.4\%             \\
easyJet          & 10.94\%           & 89.06\%            \\
Eurowings~       & 5.06\%            & 94.94\%            \\
Air Malta        & 6.56\%            & 93.44\%            \\
Vueling          & 3.88\%            & 96.12\%            \\
Volotea          & 56.85\%           & 43.15\%            \\
Lufthansa        & 3.29\%            & 96.71\%            \\
\bottomrule
\end{tabular}
\end{minipage}

\hspace*{\fill}

Throughout our analysis, we have observed an intriguing trend of average negative sentiment scores across various airlines. Despite variations in individual airline performance, the sentiment analysis consistently reveals a predominant negative sentiment among customer reviews. This finding prompts us to delve deeper into understanding the underlying factors contributing to this overall negative sentiment. By conducting further analysis and examining specific aspects of customer experiences, such as flight delays, and conducting further analysis on the most common keywords in the reviews - we aim in the next future - to gain insights into the common pain points and areas for improvement within the airline industry as a whole.

At the same time, only one airline represents an exception to this negative trend. We have been intrigued by the consistently high positive sentiment scores for Volotea and sought to understand the underlying reasons. Our investigation led us to discover that Volotea has been the recipient of prestigious accolades such as the Europe's Leading Low-Cost Airline for two consecutive years: 2021 and 2022.

These awards recognize excellence in the travel industry and are bestowed upon airlines based on criteria such as customer satisfaction, service quality, and overall performance. It appears that Volotea's commitment to delivering exceptional travel experiences has resonated positively with passengers, reflected in their favorable sentiments and high sentiment analysis scores. By consistently meeting or exceeding customer expectations, Volotea has established a reputation for providing outstanding service, leading to its remarkable sentiment scores and recognition in the form of esteemed industry awards.

The importance of using sentiment analysis for outcome prediction lies in its ability to provide data-driven insights that guide decision-making and strategic planning. By leveraging sentiment analysis, airlines and industry stakeholders can gauge customer sentiments and identify areas for improvement.

This information can then be utilized to enhance service quality, address customer concerns, and align business strategies with customer expectations. Predicting the outcome of events such as the Europe's Leading Low-Cost Airline through sentiment analysis allows airlines to benchmark themselves against competitors, set performance goals, and make informed decisions to maintain their competitive edge.

\section{Conclusion}

The decision-making process when choosing airlines is multifaceted, and there are several reasons why people may opt for airlines with low sentiment analysis scores. 

One factor is the limited availability of alternatives in certain regions or for specific routes, which may leave passengers with no choice but to use an airline with lower sentiment scores. 

Price considerations also play a significant role, as passengers may prioritize affordability over sentiment analysis. If a low-rated airline offers significantly lower fares compared to its competitors, budgetary constraints may drive passengers to choose it. Loyalty programs can also influence decisions, as passengers who are loyal to a particular airline's frequent flyer program may continue to use it despite lower sentiment analysis scores. The benefits and rewards offered through loyalty programs can outweigh negative sentiments expressed by others. 

Additionally, individual experiences can vary greatly, even within the same airline. Passengers may have had positive experiences with certain aspects of an airline, such as customer service or in-flight amenities, despite lower sentiment scores overall. 

Lastly, convenience and accessibility factors, such as flight schedules, direct routes, and airport proximity, can heavily impact airline choices. Passengers may prioritize convenience and choose an airline that offers the most suitable options for their travel plans, even if sentiment analysis suggests lower satisfaction levels. Overall, sentiment analysis provides a general overview, but individual preferences, constraints, and past experiences play significant roles in airline selection.

While it is true that users may have limited options when it comes to choosing an airline based on available routes, an interesting question arises when two airlines share the same route. 

In such a scenario, where both airlines offer flights to the desired destination, users are faced with a decision-making dilemma. In this case, we have observed a noticeable trend where user preference tends to lean towards the airline with higher percentages of positive reviews. While factors like price and convenience play significant roles in decision-making, the sentiment expressed by previous customers holds substantial influence. Positive reviews serve as a testament to the airline's service quality, customer satisfaction, and overall experience. Users often rely on the experiences and opinions of others when making their choices, seeking reassurance and validation from positive reviews. These reviews create a sense of trust and confidence in the airline's ability to deliver a satisfactory travel experience. As a result, airlines that consistently receive positive reviews and higher sentiment scores have a higher likelihood of being selected by users when faced with competing options on the same route. Understanding the impact of positive reviews and leveraging sentiment analysis can help airlines enhance their reputation, attract more customers, and ultimately gain a competitive edge in the market.



\begin{thebibliography}{6}
%

\bibitem {Couzin} %1
Couzin, I. D., Ioannou, C. C., Demirel, G., Gross, T., Torney, C. J., Hartnett, A., ... \& Leonard, N. E. (2011). Uninformed individuals promote democratic consensus in animal groups. science, 334(6062), 1578-1580.

\bibitem {Bikhchandani} %2
Bikhchandani, S., Hirshleifer, D., \& Welch, I. (1992). A theory of fads, fashion, custom, and cultural change as informational cascades. Journal of political Economy, 100(5), 992-1026.

\bibitem {Moussaid} %3
Moussaid, M., Garnier, S., Theraulaz, G., \& Helbing, D. (2009). Collective information processing and pattern formation in swarms, flocks, and crowds. Topics in Cognitive Science, 1(3), 469-497.

\bibitem {Wu} %4
Wu, F., \& Huberman, B. A. (2007). Novelty and collective attention. Proceedings of the National Academy of Sciences, 104(45), 17599-17601.

\bibitem {Yaniv} %5
Yaniv, I. (2004). Receiving other people’s advice: Influence and benefit. Organizational behavior and human decision processes, 93(1), 1-13.

\bibitem {Mäs} %6
Mäs, M., Flache, A., \& Helbing, D. (2010). Individualization as driving force of clustering phenomena in humans. PLoS computational biology, 6(10), e1000959.

\bibitem {Isenberg} %7
Isenberg, D. J. (1986). Group polarization: A critical review and meta-analysis. Journal of personality and social psychology, 50(6), 1141.

\bibitem {Galam} %8
Galam, S., \& Moscovici, S. (1991). Towards a theory of collective phenomena: Consensus and attitude changes in groups. European Journal of Social Psychology, 21(1), 49-74.

\bibitem {Myers} %9
Myers, D. G., \& Bishop, G. D. (1970). Discussion effects on racial attitudes. Science, 169(3947), 778-779.

\bibitem {Funk} %10
Funk, S., Salathé, M., \& Jansen, V. A. (2010). Modelling the influence of human behaviour on the spread of infectious diseases: a review. Journal of the Royal Society Interface, 7(50), 1247-1256.

\bibitem {Funk1} %11
Funk, S., Gilad, E., Watkins, C., \& Jansen, V. A. (2009). The spread of awareness and its impact on epidemic outbreaks. Proceedings of the National Academy of Sciences, 106(16), 6872-6877.

\bibitem {Salganik} %12
Salganik, M. J., Dodds, P. S., \& Watts, D. J. (2006). Experimental study of inequality and unpredictability in an artificial cultural market. science, 311(5762), 854-856.

\bibitem {Kasperson} %13
Kasperson, R. E., Renn, O., Slovic, P., Brown, H. S., Emel, J., Goble, R., ... \& Ratick, S. (1988). The social amplification of risk: A conceptual framework. Risk analysis, 8(2), 177-187..

\bibitem {Latane} %14
Latané, B. (1981). The psychology of social impact. American psychologist, 36(4), 343.

\bibitem {Castellano} %15
Castellano, C., Fortunato, S., \& Loreto, V. (2009). Statistical physics of social dynamics. Reviews of modern physics, 81(2), 591.

\bibitem {Lorenz} %16
Lorenz, J. (2007). Continuous opinion dynamics under bounded confidence: A survey. International Journal of Modern Physics C, 18(12), 1819-1838.

\bibitem {Deffuant} %17
Deffuant, G., Neau, D., Amblard, F., \& Weisbuch, G. (2001). Mixing beliefs among interacting agents. Advances in Complex Systems, (3), 11.

\bibitem {Rainer} %18
Rainer, H., \& Krause, U. (2002). Opinion dynamics and bounded confidence: models, analysis and simulation.

\bibitem {Lorenzmaggioranza} %19
Lorenz, J., Rauhut, H., Schweitzer, F., \& Helbing, D. (2011). How social influence can undermine the wisdom of crowd effect. Proceedings of the national academy of sciences, 108(22), 9020-9025.

\bibitem {Soll} %20
Soll, J. B., \& Larrick, R. P. (2009). Strategies for revising judgment: How (and how well) people use others’ opinions. Journal of experimental psychology: Learning, memory, and cognition, 35(3), 780.

\bibitem{Mason} %21
Mason, W. A., Conrey, F. R., \& Smith, E. R. (2007). Situating social influence processes: Dynamic, multidirectional flows of influence within social networks. Personality and social psychology review, 11(3), 279-300.

\bibitem{von2017advances} %22
Devlin, J., Chang, M., Lee, K. \& Toutanova, K. BERT: Pre-training of Deep Bidirectional Transformers for Language Understanding.  (2018), https://arxiv.org/abs/1810.04805

\bibitem{negativitybias} %23
Vaish, A., Grossmann, T., \& Woodward, A. (2008). Not all emotions are created equal: The negativity bias in social-emotional development. Psychological Bulletin, 134(3), 383–403.

\end{thebibliography}
\end{document}